\documentstyle[12pt,mkstyle,twoside]{article}
\title{Dempster-Shafer Belief Function - A New     Interpretation}
\shorttitle{DS Belief Function - A New Interpretation}

\newcommand{\rDef}[1]{Def.\ref{#1}}

\newcommand{\Bem}[1]{}
\newcommand{\Bemerkung}[1]{}

\date{}
\font\gh=eufm10 scaled \magstep1

\newcommand{\Prob}[2]{{ {\mbox{\gh Prob} ^{#2(#1)}} 
                         \atop {_{#1}} 
                     }}

\begin{document}

\machetitel
 
\section{Introduction}

Dempster-Shafer theory of evidence has been found
by many researchers 
 very attractive as a way of 
modeling reasoning behaviour under uncertainty stemming from ignorance. It 
provides a framework for representation of certainty of a logical formula 
without necessity of expressing commitment to any of its consequences. E.g. 
we can express our 100 \% belief in fact that Tweedy's wife is either Mary or 
Jane and at the same time express our total ignorance to the fact who of 
them is actually his wife (zero belief attached to the statement "Mary is 
Tweedy's wife" and zero belief in "Jane is Tweedy's wife"). 

If a theory is to become of practical importance in expert systems 
application - as foundation for knowledge representation and reasoning, ar 
least the following conditions must be fulfilled:
\begin{itemize} 
\item
there must exist an efficient method for reasoning within this framework
\item
there must exist a clear correspondence between the contents of the knowledge 
base and the real world
\item
there must be a clear correspondence between the reasoning method and some 
real world process
\item
there must exist a clear correspondence between the results of the reasoning 
process and the results of the real world process corresponding to the 
reasoning process.
\end{itemize}

Only under such circumstances we can say that the expert system is helpful as 
it allows us either to predict or to follow retrospectively real world 
processes.

Dempster initiated the theory of evidence in his paper \cite{Dempster:67} and 
other works, and Shafer developed this theory in his book \cite{Shafer:76} 
and other publications. Though it became obvious that the DST 
(Dempster-Shafer Theory) captures many intuitions behind the human dealing 
with uncertainty (e.g. as mentioned above), it did not become a foundation 
for implementation of expert systems with uncertainty due to claimed high 
 computational complexity \cite{Grzymala:91}. 
 In the recent years, however, a 
number of efficient 
methods for dealing with DS reasoning have been developed - see e.g. 
\cite{Shenoy:90} and citations therein. So the first of the  above 
mentioned 
conditions is met. Meeting of other conditions proved to be more complicated. 

Smets \cite{Smets:92} and also initially Shafer \cite{Shafer:76} insisted on 
Bels (measures of uncertainty in 
the DST) not being connected to any empirical measure (frequency, probability 
etc.) considering the domain of DST applications  as the one where 
"we are ignorant of the existence of probabilities", and 
warn that the DST is 
"not a 
model for poorly known probabilities" (\cite{Smets:92}, p.324). 
The question may be raised, however, what practically useful can be obtained 
from a computer reasoning on the basis of such a DST. It would have to be 
demonstrated that humans indeed reason as DST. Then the computer, if fed 
with our knowledge, would 
be capable to predict our conclusions on a given subject. However, to my 
knowledge, no experiment
confirming that humans actually use internally DST for reasoning under
uncertainty
 has been carried out. Under these circumstances 
the computer reasoning with DST would tell us what we have to think and 
not 
what we think. Hence, from the point of view of computer implementation the 
position of Smets and Shafer is not acceptable.

The other category of DST interpretations, described by Smets as approaches 
assuming existence of an underlying probability distribution, which is only 
approximated by the Bels (called by him 
PXMY models), is represented  by early works of Dempster \cite{Dempster:67}, 
papers 
of Kyburg \cite{Kyburg:87}, Fagin \cite{Fagin:91}, 
\cite{Fagin:91B},, Halpern \cite{Halpern:92}, 
Skowron \cite{Skowron:93}, Grzyma{\l}a-Busse \cite{Grzymala:91} and others. 
Both 
Smets \cite{Smets:92} and 
Shafer \cite{Shafer:90b} consider such approaches as inadequate as most of 
them give rise 
to contradictions and counter intuitiveness. As Smets states,
"Far too often, authors concentrate on the static component (how beliefs are 
 allocated?) and discover many relations between TBM (transferable belief 
model 
of Smets) and ULP (upper lower probability) models, inner and outer measures 
(Fagin and Halpern \cite{Fagin:89}), random sets (Nguyen \cite{Nguyen:78}), 
probabilities of provability 
 (Pearl \cite{Pearl:88}), probabilities of necessity (Ruspini 
\cite{Ruspini:86}) etc. But these authors 
usually do not explain or justify the dynamic component (how are beliefs 
updated?), that  is, how updating (conditioning) is to be handled (except in 
some cases by defining conditioning as a special case of combination). So I 
 (that is Smets) feel that these partial comparisons are incomplete, 
especially 
as all these interpretations lead to different updating rules." 
(\cite{Smets:92}, pp. 324-325). Smets attributes this failure to the very 
nature of attempts  of assigning a probabilistic interpretation. We disagree 
with Smets and will show in this paper that creation of a probabilistic 
interpretation of the DST incorporating the Dempster rule of combination is 
actually possible.  However, this new interpretation indicates the need for 
a drastic change  in  viewing  the  Dempster  rule:  it  does  not 
accommodate 
evidence, but prejudices. How this statement is to be understood, will be 
visible later. Nonetheless our interpretation allows for assignment of an 
experimentally verifiable numerical meaning to a DS knowledge base, assigns a 
numerical meaning to the reasoning process (the DS rule of combination) and 
yields agreement between numerical empirical 
interpretation of the results of DS reasoning and results of a real world 
process. This means that we have an interpretation fitting formal 
interpretation of the DS theory to the largest extent ever achieved. 

Smets (\cite{Smets:92},p.327) subdivided  the DST into two categories: a 
closed 
world category (as if excluding the possibility of contradictions in the 
"evidence") and an open world category of DST (as if allowing for this). Let 
us assume that two independent experts elicited their beliefs concerning 
the event A: both assigned beliefs of 0.7 to the event A, and beliefs of 0.3 
to the event $\lnot$A. The open world DST will lead us to a combined belief 
in A of 0.5 and in  $\lnot$A of 0.1. The closed world assumption on the other 
hand will assign   a combined belief 
in A of 0.7 and in  $\lnot$A of 0.3. I find it a dismaying property of a  
theory if collecting agreeing information from independent expert shall 
decline my belief in the opinions of both experts. Hence only closed world 
theories are subject of this paper.

We first recall the formal definition of the DS-Theory,
then introduce some notation used throughout the rest of the paper. 
Subsequently 
we develop our interpretation of the joint belief distribution and of  
evidential updating. We conclude with a brief comparison 
of our interpretation with other attempts.\\

\section{Formal  Definition  of  the  Dempster-Shafer  Theory   of 
Evidence}

Let us make the remark that if an object is described by a set of 
discrete attributes $X_1,X_2,...,X_n$ taking  values  from  their 
respective domains $\Xi_1,\Xi_2,...,\Xi_n$  then we can 
think of it  as 
being described by a complex attribute X  having  vector  values, 
that is the domain $\Xi$ of X is equal:
$$\Xi=\{(x_1,x_2,...,x_n) | x_i \in \Xi_i , i=1,...,n\}$$.

So unless specified otherwise let us assume that we  are  talking 
of objects described by a single attribute X  taking  its  values 
from the domain $\Xi$. We say that $\Xi$, the domain of X
 is our space of discourse 
spanned by the attribute X. We shall also briefly say that X is our
space of discourse instead.

For the purpose of  this  paper  we  define  the  Bel-function  as 
follows (compare also \cite{Halpern:92}, \cite{Smets:92},
 \cite{Shafer:90b}):

\begin{df} \label{BelDef}
  The Belief Function in the sense of the DS-Theory is defined 
as Bel:$2^\Xi \rightarrow [0,1]$ with 
$\Xi=\Xi_1 \times Xi_2 \times ... \times \Xi_n$ being the 
space spanned by the  attribute $X= X_1 \times X_2 \times \dots \times 
X_n$ with 
 $$\forall_{A;A \subseteq \Xi} \quad Bel(A) = \sum_{B \subseteq A} m(B)$$
where m(A) is a Mass Function in the sense of the DS-Theory 
(see \rDef{mDef} below). 
\end{df}
 
The function m is defined as 
\begin{df} \label{mDef}
The Mass Function in the sense of the DS-Theory is defined as
m:$2^\Xi \rightarrow [0,1]$
with
$$  m(\emptyset)=0$$
$$  \sum_{A \in 2^\Xi} m(A)=1 $$
$$  \forall_{ A \in 2^\Xi } \quad m(A) \geq 0 $$.
\end{df}
\begin{df}
 Whenever $m(A) > 0$, we say that A is the focal  point  of  the 
Bel-Function.
\end{df}

Let  us  also  introduce  the  Pl-Function   (Plausibility) as:
\begin{df} The Plausibility Function in the sense of the DS-Theory is 
defined as 
$$\forall_{A; A \subseteq \Xi} \quad Pl(A) = 1-Bel(\Xi-A ) $$
\end{df}

Beside the above  definition  a  characteristic  feature  of  the 
DS-Theory is the so-called DS-rule of combination of  independent 
evidence:
\begin{df}
   Let   $Bel_{E_1}$    and    $Bel_{E_2}$     represent 
independent information over the same space of discourse. Then:
    $$Bel_{E_1,E_2}=Bel_{E_1} \oplus Bel_{E_2}$$ 
defined as:
$$m_{E_1,E_2}(A)=c \cdot  \sum_{B,C; A= B \cap C} m_{E_1}(B) \cdot 
m_{E_2}(C)$$ 
(c - normalizing constant) represents the Combined Belief-Function  of 
Two Independent Beliefs
\end{df}

\section{Denotation}

F. Bacchus in his paper \cite{Bacchus:90} on axiomatization of 
probability theory and 
first order logic shows that probability should be considered as a quantifier 
binding free variables
in first order logic expressions just like universal and existential 
quantifiers do. So if e.g. $\alpha(x)$ is an open expression with a free 
variable $x$ then   $[\alpha(x)]_x$ means the probability of truth of the 
expression  $\alpha(x)$. 
(The quantifier $[]_x$ binds the free variable $x$ and yields a numerical 
value ranging from 0 to 1 and meeting all the Kolmogoroff axioms). 
Within the expression  $[\alpha(x)]_x$ the variable 
$x$ is bound. See \cite{Bacchus:90} on justification why other types of 
 integration of probability theory and first order logic or propositional 
logic fail. Also for justification of rejection
of the traditional view of probability as a function over sets.
While sharing Bacchus' view, we find his notation a 
bit cumbersome so we change it to be similar to the universal and 
existential quantifiers throughout this paper.
Furthermore, Morgan \cite{Morgan:91} insisted that the probabilities be 
always considered in close connection with the population they refer to.  
 Bacchus' expression 
$[\alpha(x)]_x$ we rewrite as:\\
  $\Prob{x}{P}\alpha(x)$ - the probability of  $\alpha(x)]$ being true within 
the population P. The  P (population) is a unary predicate with P(x)=TRUE 
indicating that the object x($\in \Omega$, that is element of a universe of 
objects) belongs to the population under considerations. If P and P' are 
populations such that $\forall_x P'(x)\rightarrow P(x)$ (that is membership 
in P' implies membership in P, or in other words: P' is a subpopulation of 
P), then we distinguish two cases:\\
case 1: $(\Prob{x}{P}P'(x))=0$ (that is probability of membership in P' with 
respect to P is equal 0) - then (according to \cite{Morgan:91} for any 
expression $\alpha(x)$ in free variable x the following holds for the 
population P': $(\Prob{x}{P'}\alpha(x))=1$\\
case 2: $(\Prob{x}{P}P'(x))>0$then (according to \cite{Morgan:91} for any 
expression $\alpha(x)$ in free variable x the following holds for the 
population P': 
$$(\Prob{x}{P'}\alpha(x))=  \frac {\Prob{x}{P}(\alpha(x) \land P'(x))}
                                {\Prob{x}{P}P'(x)}$$  
We also use the following (now traditional) mathematical symbols:\\
$\forall_{x}\alpha(x)$ - always  $\alpha(x)$ (universal quantifier) \\
$\exists_{x}\alpha(x)$ - there exists an x such that $\alpha(x)$ 
(existential quantifier) \\
\begin{tabular}{lp{7cm}}
$\alpha \land \beta$ & - logical AND of expressions\\
$\bigwedge_{B} \alpha(B)$  & - logical AND over all  instantiations of
the expression $\alpha(B)$ in free 
variable $B$\\
$\alpha \lor \beta$  & - logical OR of expressions\\
$\bigvee_{B} \alpha(B)$  & - logical OR over all  instantiations of
the expression $\alpha(B)$ in free 
variable $B$\\
$\lnot$  & - logical negation\\
$P \cap Q$  & - intersection of two sets\\
$P \cup Q$  & - union of two sets\\
\end{tabular}

\section{A New Interpretation of Belief Functions}

The empirical meaning of a
new interpretation of the DS Belief function will be explained
by means of the following example:

\begin{Bsp} \label{COOTexample}
Let us consider a daily-life example. Buying a bottle of hair shampoo is 
not a trivial task from both the side of the consumer and the 
manufacturer. If the consumer arrives at the consciousness that the shampoos 
may fall into one of the four categories: 
high quality products 
(excellent for maintaining cleanness and health of the consumer)
 (H), 
moderate quality products (keeping just all Polish industry standards) (M), 
suspicious products  (violating some industry standards) (S)
 and products dangerous for health and life (containing bacteria or fungi or 
other microbes causing infectious or invasive diseases, containing 
cancerogenous or poisonous substances etc.) (D), 
he has a hard time upon leaving his house for shopping. Clearly, precise 
chemical, biochemical and medical tests exist which may precisely place
the product into one of those obviously exclusive categories. But the
 Citizen\footnote{The term "Citizen" was a fine socialist time descriptor 
allowing to avoid the cumbersome usage of words like "Mr.", "Mrs."  and 
 "Miss"} Coot\footnote{This family name was coined as abbreviation for 
"Citizen 
Of Our Town"} usually neither has a private chemical laboratory nor enough 
money to make use of required services. Hence Citizen Coot coins a personal 
set of "quality" tests $M ^1$ mapping the pair (bottle of shampoo, quality) 
into the set \{TRUE, FALSE\} (the letter O - object - stands for bottle of 
shampoo, H, M, S, D indicate quality classes: high, moderate, suspicious, 
dangerous): \\
\begin{enumerate}
\item If the shampoo is heavily advertised on TV then it is of 
high quality ($M ^1(O,\{H\})=TRUE$) and otherwise not ($M ^1(O,\{H\})=FALSE$).
\item If the name of the shampoo was never heard on TV, 
 but the 
bottle looks fine 
(pretty colours, aesthetic shape of the bottle), then the shampoo must be of 
moderate quality  ($M ^1(O,\{M\})=TRUE$)  and otherwise not 
($M ^1(O,\{M\})=FALSE$).
\item   If  the packaging is not fine or  the date of production is not 
readable on the bottle or 
the product is out of date, but the shampoo  smells acceptably 
otherwise 
 then it is suspicious  ($M ^1(O,\{S\})=TRUE$)  and 
otherwise not ($M ^1(O,\{S\})=FALSE$).
\item  If either the packaging is not fine or 
the date of production is not readable on the bottle or 
the product is out of date,
and at the same time 
the shampoo smells awfully,  then it is dangerous  
($M ^1(O,\{D\})=TRUE$  and otherwise not ($M ^1(O,\{D\})=FALSE$).
\end{enumerate} 

Notice that the criteria are partially rational: a not fine looking bottle 
may in fact indicate some decaying processing of the shampoo or at least that 
the product remains for a longer time on the shelf already. Bad smell is 
usually caused by development of some bacteria dangerous for human health.
Notice also that test for high and moderate quality are enthusiastic, while 
the other two are more cautious. 

Notice that the two latter tests are more difficult to carry out in a shop
than the leading two (the shop assistant would hardly allow to open a  
bottle before buying). 
Also, there may be no time to check whether the shampoo was actually
advertised on TV or not (as the son who carefully watches 
all the running advertisements stayed home and does his lessons).
Hence some simplified tests may be quite helpful:
\begin{itemize} 
\item $M ^1(O,\{S,D\})$: If  the packaging is not fine or the product is out 
of date or the production date is not readable  then the product is 
either suspicious or dangerous  ($M ^1(O,\{S,D\})=TRUE$  and otherwise not 
($M ^1(O,\{D,S\})=FALSE$).
 .
\item $M ^1(O,\{H,M\})$: If the packaging looks fine, then the product is 
either of high    or moderate quality  ($M ^1(O,\{M,H\})=TRUE$  and otherwise 
not ($M ^1(O,\{M,H\})=FALSE$)..
\end{itemize}
 Clearly these  tests are far from being precise ones, but for the Citizen 
Coot 
no better tests will be ever available. What is more, they are not exclusive: 
if one visits a dubious shop at a later hour, one may buy a product meeting 
both  $M ^1(O,\{H\})$ and  $M ^1(O,\{D\})$ as defined above ! 

Let us assume we have two types of shops in our town: good ones (G) and bad 
ones (B). (Let $M ^2:\Omega \times 2^\{G,B\} \rightarrow \{TRUE, FALSE\}$ 
indicate for each shampoo in which shop type it was available. Further, let  
$M ^3:\Omega \times 2^{\{ H, M, S, D \} \times \{G,B\}} \rightarrow \{TRUE, 
FALSE\}$ indicate for each shampoo both its quality and the type of shop it 
was available from. Let clearly $M ^1(O,Quality) \land M ^2(O,Shop)=M 
^3(O,Quality \times Shop)$.\\
The good shops are those  with  new  furniture,  well-clothed  shop 
assistants. Bad 
ones are those with always dirty floor or old furniture, or badly clothed 
shop assistants. Clearly, again, both shop categories may be considered 
(nearly) exclusive as seldom well clothed shop assistants 
 do not care of floors. Let us assume we have obtained the 
statistics of shampoo sales in our town presented in Table \ref{statshamp}:\\

\begin{table} \label{statshamp}
\caption{Sold shampoos statistics}
\begin{center}
\begin{tabular}{r|rrr|r}
      Quality true for & Shop type  B &  G  & B,G & Total\\
\hline
          H &     20   & 100 &  70 & 190  \\
          M &     80   & 100 & 110 & 290      \\
          S &     50   &   5 &  15 &  70      \\
          D &     10   &   1 &   3 &  14     \\
        H,S &     15   &  10 &  14 &  39      \\
        M,S &     30   &  20 &  25 &  75      \\
        H,D &      8   &   2 &   3 &  13     \\
        M,D &     15   &   7 &  10 &  32      \\
\hline
      total &    228   & 245 & 250 & 723      \\
\end{tabular}
\end{center}
\end{table}
%
Rows and columns are marked with those singleton tests which were passed
(e.g. in the left upper corner there are 20 
shampoo bottles sold in an undoubtedly bad shop and having exclusively
high quality, that is for all those bottles (O)
$M ^1(O,\{H\})=TRUE$, $M ^1(O,\{M\})=FALSE$,  $M ^1(O,\{S\})=FALSE$, $M 
^1(O,\{D\})=FALSE$, and  $M ^2(O,\{B\})=TRUE$,  $M ^2(O,\{G\})=FALSE$.)
The measurement of $M ^1(O,\{H\})$ would yield TRUE for 190+39+13  =242 
bottles and 
FALSE for the remaining 581 bottles, the measurement of $M ^1(O,\{D\})$ would 
yield 
TRUE for 14+13+32=59 bottles, and FALSE for the remaining 664 bottles. The 
 measurement $M ^1(O,\{S,D\})$ will turn true in 70+14+ 39+75+ 13+12 =343 
cases and FALSE in the remaining 480 cases.
\end{Bsp}

In general let
us assume that we know that objects of a  population  can 
be described by an  intrinsic attribute  X taking 
exclusively   one   of   the   n   discrete   values   from   its 
domain $\Xi=\{v_1,v_2,...,v_n\}$ . Let us  assume  furthermore 
that to obtain knowledge of the actual value taken by  an  object 
we must apply a measurement method (a system of tests) $M$  

\begin{df} \label{MDef}
$X$ be a set-valued attribute taking as its values non-empty
subsets of a finite domain $\Xi$.
By a measurement method 
of value of the attribute $X$
we understand a function:
 $$M: \Omega \times 2^\Xi \rightarrow \{TRUE,FALSE\}$$. 
where $\Omega$ is the set of objects, (or population of objects)
such that 
\begin{itemize}
\item
 $ \forall_{\omega; \omega \in \Omega} \quad
M(\omega,\Xi)=TRUE$ (X takes at least one of values from $\Xi$)
\item
 $ \forall_{\omega; \omega \in \Omega} \quad
M(\omega,\emptyset)=FALSE$ 
\item 
 whenever 
$M(\omega,A)=TRUE$
for $\omega \in \Omega$, $A \subseteq \Xi$
 then for any $B$ such that $A \subset B$ $M(\omega,B)=TRUE$   
holds,
 \item 
 whenever 
$M(\omega,A)=TRUE$
for $\omega \in \Omega$, $A \subseteq \Xi$ and if $card(A)>1$ then there 
exists  $B$, $B \subset A$ such that $M(\omega,B)=TRUE$ holds.
\item 
for every $\omega$ and every $A$
either  
$M(\omega,A)=TRUE$  or 
 $M(\omega,A)=FALSE$ (but never both).
 \end{itemize}
$M(\omega,A)$  tells us whether or not any of the elements of the set A 
belong to the actual value of the attribute $X$ for the object $\omega$.
 \end{df}

The measuring function M(O,A), if it takes the value TRUE,  
states for an object O and a set A of values from the domain of X  
that  the X 
takes for this object (at least) one of the  values  in A. 

It makes sense to talk of such measuring function assigning truth values 
to sets of values of an attribute if it is possibly cheaper to measure   
M(O,A) 
 than to measure M(O,B) whenever $B \subset A$ and we are interested in 
avoiding
measuring M(O,B) whenever possible, that is whenever measuring M(O,A) 
suffices.  For example, measuring pH-value with a pH-meter may turn out to be 
more expensive than one with litmus paper, at the advantage of a higher 
precision.  

The above definition assumes that   this    
measurement method is superset- and subset-consistent that is: Whenever 
$M(object,A)=TRUE$ then
  $$\forall_{B; A \subset B} \quad M(object,B)=TRUE$$  
 holds, and if $card(A)>1$ then
 $$\exists_{B; B \subset A} \quad M(object,B)=TRUE$$
 holds. The superset consistency means that if a test for 
larger set of values indicates FALSE then it is not necessary to test its 
subsets as they will not contribute to our knowledge of the  value 
of X (cost 
savings). The subset consistency means that if the M-test for a given value 
set gives true than in end effect at least of its singleton subsets would 
yield TRUE for the respective M-test. It is clearly the matter of convention: 
we assume that we can always provide the answer YES or NO, and whenever 
we are in doubt we still answer YES.   

Such a convention is not an unusual one: in various legal 
systems "anything, 
that is not forbidden by law, is permitted"; in the default logics if a 
default statement cannot be proven wrong, it is assumed correct. 

In any case, this means that from the universe of all possible objects, a 
concrete      measurement method selects a population for which its 
assumptions are 
satisfied. E.g. if we have a measurement method for measuring pH-values, 
we surely consider an aqueous sodium solution as a member of our universal 
population, but never a car as such (because then pH-value has no meaning 
at all)..

Furthermore us consider this measurement method a stable  one 
that is whenever the same object is presented,  the  results  are 
the same.
However, let  us  assume  that  the  measurement  method  is  not 
completely reliable: it measures only 
quantities related to the
 quantity  X 
and not X itself. So it is conceivable that e.g.
$M(object,\{v_1\})=TRUE$    and     at     the     same     time 
$M(object,\{v_2\})=TRUE$  though both values of X are
deemed to be 
 exclusive. 
For practical reasons however it may not  bother  us  at  all  as 
either the true value of X may not be accessible at all (e.g. the 
true event of killing or not killing  a  person  by  the  suspect 
belongs to the past and can never be recalled as  such),  may  be 
too expensive to access (e.g. if  the  most  reliable  method  of 
checking whether a match can inflame or not it to inflame it, but 
thereafter it would be useless, so we check only for  its  color, 
dryness etc.) or it may be prohibitive to  access  it  for  other 
reasons, e.g. social (sex may  be  treated  with  extremely  high 
precision as an exclusive attribute  taking  values  male,female, 
but we  would  reluctantly  check  the  primary  features  before 
deciding to call someone Mr, Miss or Mrs). 
Beside  this  it  may  prove  too  expensive  to  check  all  the 
elementary hypotheses (e.g. in the  medical  diagnosis)  so  that 
after stating $M(object,\{v_1\})=TRUE$ we do not bother of other 
alternatives, that is of the degree of imprecision of the 
relationship between the measured quantities and the real values of X.
We assume that the approximations of X  achieved by the measurement method
are in most cases sufficient for our decision making (whatever its nature),
so we do not insist on closer knowledge of X itself.

So though we wish $X$ to take singleton values only, we actually live with the 
fact that for our practical purposes $X$ is possibly set-valued.\\

     Let  us  make  at  this  point  some  remarks  on  practical 
relevance. 
\begin{Bsp}
If we are making  statistical  tests  on  equality  or 
non-equality of two quantities (means, variances, distributions), 
we can purely logically say that the quantities are either  equal 
or not equal but never  both.  However,  the  available  indirect 
measurement method (by sampling) may lead  to  a  statement  that 
there is neither  evidence  to  reject  equality  nor  to  reject 
non-equality. So we say  that  in  those  cases  both  equity  and 
inequity 
 holds. We still enjoy statistical inference because in sufficiently many 
other cases statistics provides us with more precise results.
 \end{Bsp}

\begin{Bsp}
 Similarly if we consider components of a chemical 
substance, the measurement methods for absence and presence of  a 
component may be different from one another depending whether  or 
not we should be more sensitive to its presence  or  absence  and 
hence  in  some  cases  applying  both  may  lead  to  apparently 
contradicting results. 
\end{Bsp}

     Let us furthermore assume that with each application of  the 
measurement  procedure  some  costs  are  connected,   increasing 
roughly with the decreasing size of the tested set A so that  we 
are ready to accept results of previous measurements in the  form 
of pre-labeling of the population. So 

\begin{df}
A {\em label} $L$ of an object $\omega \in \Omega$ is a subset of the domain
$\Xi$ of the attribute $X$. \\
A {\em labeling}  under the measurement method $M$  is a function $l: \Omega 
\rightarrow 2^\Xi$ such that for any object  $\omega \in \Omega$ either
$l(\omega)=\emptyset$ or $M(\omega,l(\omega))=TRUE$.\\
Each {\em labelled object}  (under the labeling $l$) 
consists of a 
pair $(O_j,L_j)$, $O_j$ - the j$^{th}$ object, $L_j=l(O_j)$ - its label.\\
By a {\em population  under the labeling $l$} we understand the predicate 
$P:\Omega \rightarrow \{TRUE,FALSE\}$ of the form 
$P(\omega)=TRUE \  iff \ l(\omega) \neq \emptyset$
(or alternatively, the set of objects  for which this predicate is true) \\
 If for every  object of the 
population the label is equal 
 to $\Xi$ then  we  talk  of  an  {\em unlabeled  population} (under the 
labeling $l$), otherwise of a {\em pre-labelled} one.
\end{df}

     Let  us  assume  that  in  practice  we  apply  a   modified 
measurement method 
$M_l$ being a function:

\begin{df} 
Let $l$ be a labeling under the measurement method $M$. 
Let us consider the population under this labeling.
The modified measurement method 
$$M_l:
 \Omega \times 2^\Xi \rightarrow 
\{TRUE,FALSE\}$$
where $\Omega$ is the set of objects, 
is is defined as  
$$M_l(\omega,A)= M(\omega,A \cap l(\omega) )$$  
(Notice that 
$M_l(\omega,A)=FALSE$ whenever $A \cap l(\omega)= \emptyset$.)
\end{df}

For a labeled object $(O_j,L_j)$  ($O_j$ - proper object, 
$L_j$  - 
its label)  and a set A of values from the domain of X, 
the modified measurement method tells us 
that $X$ takes one of the values in A if and only if it takes in fact 
a value from intersection of A and $L_j$.
 Expressed   differently,   we 
discard a priori any attribute not in the label.

Please pay attention also to the fact, that given a population P for which 
the measurement method $M$ is defined, the labeling $l$ (according to its 
definition) selects a subset of this population, possibly a proper subset, 
namely the population P'
under this labeling. 
$P'(\omega)=P(\omega) \land M(\omega,l(\omega))$. 
Hence also $M_l$ is defined possibly for the "smaller" 
population P' than $M$ is. \\

\begin{Bsp}
     In practice, we  frequently  have  to  do  with  pre-labelled 
population. The statistics of illnesses  based  on  poly-clinical 
data are based on a population pre-labelled by  financial  status 
(whether or not they are ready to visit  a  physician  with  less 
serious  disease  due  to  economical  background),   educational 
background (whether or not they estimate properly the seriousness 
of the disease, whether  or  not  they  care  of  symptoms)  etc. 
Similarly in chemical analysis knowledge of substrates pre-labels 
the  tests  on  composition  of   the   product   (not   relevant 
measurements are a priori discarded) etc.
\end{Bsp}

\begin{Bsp} To continue Citizen Coot example, we may
believe that in good shops only moderate and high quality products
are available, that is we assign to every shampoo $\omega$
the label $l(\omega)=\emptyset$
(we discard it from our register)
 if $\omega$ denies our belief
that there are no suspicious nor dangerous products in a good shop,
and $l(\omega)=\{H,M\}$ if it is moderate or high quality
product in a good shop
and
$l(\omega)=\Xi$  to all the other products. After this 
rejection of shampoos not fitting our beliefs we have to do 
with (a bit smaller) sold-shampoos-population from Table 
\ref{modstatshamp} :
\\

\begin{table} \label{modstatshamp}
\caption{Modified sold shampoos statistics}
\begin{center}
\begin{tabular}{r|rrr|r}
      Quality true for & Shop type  B &  G  & B,G & Total\\
\hline
          H &     20   & 112 &  70 & 202  \\
          M &     80   & 127 & 110 & 317      \\
          S &     65   &   0 &   0 &  65      \\
          D &     13   &   0 &   0 &  13     \\
        H,S &     15   &   0 &  14 &  29      \\
        M,S &     30   &   0 &  25 &  55      \\
        H,D &      8   &   0 &   3 &  11     \\
        M,D &     15   &   0 &  10 &  25      \\
\hline
      total &    246   & 239 & 232 & 717      \\
\end{tabular}
\end{center}
\end{table}

Please notice the following changes: Suspicious and dangerous products
encountered in good shops were totally dropped from the statistics
(their existence was not revealed to the public). Suspicious and dangerous 
products from shops with unclear classification (good/bad shops) were 
declared to come from bad shops. Products from good shops which obtained both 
 the  label high quality and dangerous were simply moved into the category 
high quality products (the bad smelt was just concealed) etc.  This is 
frequently the sense in which our beliefs have impact on our attitude towards 
real facts and we will see below that the Dempster-Shafer Theory 
reflects such a view of beliefs. 
\end{Bsp}

Let us now define the following function:

\begin{df}
$$Bel_P ^{M}(A)=\Prob{O}{P}(\lnot M(O,\Xi-A))$$
which is the probability that the test M, while being true for A, rejects 
every hypothesis of the form X=$v_i$ for every  $v_i$  not  in  A 
for the population P.
We shall call this function "the belief exactly in the the result 
of measurement". 
\end{df}

Let us define also the function:

\begin{df}
$$Pl_P ^{M}(A)=\Prob{O}{P}(  M(O,A))$$
which  is  the  probability  of  the  test  M  holding  for  
 A for the population P. Let us refer to this function as  the 
"Plausibility of taking any value from the set A".
\end{df}

Last not least be defined the function:
\begin{df}
$$m_P ^{M}(A)=\Prob{O}{P}( \bigwedge_{B;B=\{v_i\}\subseteq A} M(O,B)
  \land \bigwedge_{B;B=\{v_i\}\subseteq \Xi-A} \lnot M(O,B) ) $$
which is the probability that all the  tests  for  the  singleton 
subsets of A are true and those outside of A are  false  for  the 
population P.
\end{df}

Let us illustrate the above concepts with Citizen Coot example:

\begin{Bsp} \label{nonlabelEx}
For the belief function for sold-bottles-population  and the measurement 
function $M ^3$, if we identify probability with relative frequency, we have 
the focal points given in the Table \ref{nonlabtable}:
\end{Bsp}

\begin{table} \label{nonlabtable} 
\caption{Mass and Belief Function under Measurement Method $M ^3$}
\begin{center}
\begin{tabular}{|l|r|r|r|} \hline
Set                           &$m_P ^{M ^3}$& $Bel_P ^{M ^3}$\\
\hline
\{(H,B)                    \} &  20/723     &   20/723       \\
\{(H,G)                    \} & 100/723     &  100/723       \\
\{(H,B),(H,G)              \} &  70/723     &  190/723       \\
\{(M,B)                    \} &  80/723     &   80/723       \\
\{(M,G)                    \} & 100/723     &  100/723       \\
\{(M,B),(M,G)              \} & 110/723     &  290/723       \\
\{(S,B)                    \} &  50/723     &   50/723       \\
\{(S,G)                    \} &   5/723     &    5/723       \\
\{(S,B),(S,G)              \} &  15/723     &   70/723       \\
\{(D,B)                    \} &  10/723     &   10/723       \\
\{(D,G)                    \} &   1/723     &    1/723       \\
\{(D,B),(D,G)              \} &   3/723     &   14/723       \\
\{(H,B),(S,B)              \} &  15/723     &   85/723       \\
\{(H,G),(S,G)              \} &  10/723     &  115/723       \\
\{(H,B),(S,B),(H,G),(S,G)  \} &  14/723     &  299/723       \\
\{(M,B),(S,B)              \} &  30/723     &  160/723       \\
\{(M,G),(S,G)              \} &  20/723     &  125/723       \\
\{(M,B),(S,B),(M,G),(S,G)  \} &  25/723     &  435/723       \\
\{(H,B),(D,B)              \} &   8/723     &   38/723       \\
\{(H,G),(D,G)              \} &   2/723     &  103/723       \\
\{(H,B),(D,B),(H,G),(D,G)  \} &   3/723     &  217/723       \\
\{(M,B),(D,B)              \} &  15/723     &  105/723       \\
\{(M,G),(D,G)              \} &   7/723     &  108/723       \\
\{(M,B),(D,B),(M,G),(D,G)  \} &  10/723     &  336/723       \\
\hline 
 \end{tabular}
\end{center}
\end{table}

It is easily seen that:


\begin{th} 
$m_P ^{M}$ is the mass Function in the sense of 
DS-Theory.
\end{th}
\AnfBeweis We shall recall the definition and construction of the DNF 
(Disjunctive Normal Form). If, given an object O of a population P under the 
measurement method M, we look 
at the expression
 $$expr(A)= \bigwedge_{B;B=\{v_i\}\subseteq A} M(O,B)
  \land \bigwedge_{B;B=\{v_i\}\subseteq \Xi-A} \lnot M(O,B)$$
for two different sets $A_1,A_2 \subseteq \Xi$ then clearly 
$expr(A_1) \land expr(A_2)$ is never true - the truth of the one excludes the 
truth of the other. They represent mutually exclusive events in the sense of 
the probability theory. On the other hand:
$$ \bigvee_{A;A \subseteq \Xi} expr(A) = TRUE $$ 
hence:
$$ (\Prob{O}{P}(\bigvee_{A;A \subseteq \Xi} expr(A))) = (\Prob{O}{P} TRUE)=1 
$$ 
and  due to mutual exclusiveness:\\
$$ \sum_{A;A \subseteq \Xi} (\Prob{O}{P} expr(A)) = 1 $$ 
which means:\\
$$ \sum_{A;A \subseteq \Xi} m_P ^M(A) = 1 $$ 
Hence the first condition of \rDef{mDef} is satisfied.
Due to the second condition of \rDef{MDef} we have\\
$$(\Prob{O}{P} expr(\emptyset))=1-(\Prob{O}{P}(M(O,
\Xi)))=$$ $$=1-(\Prob{O}{P} TRUE)=1-1=0$$
Hence $$ m_P ^M(\emptyset)=0$$.
 The last condition  is 
satisfied due to the very nature of probability: Probability is never 
negative. So we can  state that $m_P ^M$ is really a Mass Function in the 
sense 
of the DS-Theory.
\EndBeweis

\begin{th} 
$Bel_P ^{M}$ is a Belief Function in the sense of 
DS-Theory   corresponding  to the  $m_P ^{M}$. 
\end{th}
\AnfBeweis 
Let A be a non-empty set.
By definition 
$$M(O,\Xi-A)= \bigvee_{C= \{v_i\} \subseteq \Xi-A} M(O,C)$$
hence by de-Morgan-law:\\
$$\lnot M(O,\Xi-A)= \bigwedge_{C= \{v_i\} \subseteq \Xi-A}\lnot M(O,C)$$
On the other hand, $\lnot M(O,\Xi-A)$ implies $M(O,A)$.\\
But :
$$M(O,A)=
\bigvee_{B \subseteq A} 
\left( \bigwedge_{C;C=\{v_i\}\subseteq B} M(O,C)
  \land \bigwedge_{C;C=\{v_i\}\subseteq A-B} \lnot M(O,C)
\right) $$
\\
So .

\noindent 
$$\lnot M(O,\Xi-A)= \lnot M(O,\Xi-A) \land M(O,A)=$$
$$=
 \bigwedge_{C;C=\{v_i\}\subseteq \Xi-A} \lnot M(O,C)
  \land M(O,A)=
$$
$$ =
 \bigwedge_{C;C=\{v_i\}\subseteq \Xi-A} \lnot M(O,C)
  \land
\left( 
\bigvee_{B \subseteq A} 
\left( \bigwedge_{C;C=\{v_i\}\subseteq B} M(O,C)   \land \right.  \right.
$$
$$
\left. \left. \land  \bigwedge_{C;C=\{v_i\}\subseteq A-B} \lnot M(O,C)
\right)  \right)
=
$$

\noindent 
$$=\bigvee_{B \subseteq A} 
\left(
 \bigwedge_{C;C=\{v_i\}\subseteq B} M(O,C)
  \land \bigwedge_{C;C=\{v_i\}\subseteq \Xi-A} \lnot M(O,C)  
\land \right.$$
$$  \left.  \land \bigwedge_{C;C=\{v_i\}\subseteq A-B} \lnot M(O,C)
\right)
=$$

\noindent 
$$=\bigvee_{B \subseteq A} 
\left( \bigwedge_{C;C=\{v_i\}\subseteq B} M(O,C)  \linebreak 
  \land \bigwedge_{C;C=\{v_i\}\subseteq \Xi-B} \lnot M(O,C) 
\right)
$$
Hence \\
$$\lnot M(O,\Xi-A) =
\bigvee_{B \subseteq A} expr(B)$$
and therefore:
$$(\Prob{O}{P} \lnot M(O,\Xi-A) ) =
(\Prob{O}{P} \bigvee_{B \subseteq A} expr(B))$$
expr(A) being defined as in the previous proof.
As we have shown in the proof of the previous theorem, expressions under the 
probabilities
of the right hand side are exclusive events, and therefore:
$$(\Prob{O}{P} \lnot M(O,\Xi-A) )=\sum_{B \subseteq A} (\Prob{O}{P} 
expr(B))$$ that is:
 $$Bel_P ^M(A \in 2^\Xi ) = \sum_{B \subseteq A} m_P ^M(B)$$
As the previous theorem shows that $m_P ^{M}$ is a DS Theory Mass 
Function, it suffices to show the above.
 \EndBeweis

\begin{th} 
$Pl_P ^{M}$ is a Plausibility Function in the sense of 
DS-Theory and it is the Plausibility  Function  corresponding  to 
the  $Bel_P ^{M}$. 
\end{th} %
 \AnfBeweis  By definition:
$$Pl_P ^{M}(A)=\Prob{O}  M(O,A)$$
hence
$$Pl_P ^{M}(A)=1-(\Prob{O}{P} \lnot M(O,A))$$
But by definition:
$$(\Prob{O}{P} \lnot M(O,A))=(\Prob{O}{P} \lnot M(O,\Xi-(\Xi-A)))=Bel_P 
^{M}(\Xi-A)$$ hence 
$$Pl_P ^{M}(A)=1-Bel_P ^{M}(\Xi-A)$$
 \EndBeweis

Two  important remarks must be made concerning this particular 
interpretation:
\begin{itemize}
\item 
Bel and Pl are both defined, contrary to many traditional approaches, as THE 
probabilities and NOT as lower or upper bounds to any probability.
\item
It is Pl(A) (and not Bel(A) as assumed traditionally) that expresses the 
probability of A, and Bel(A) refers to the probability of the complementary 
set $A ^C$.\\
\end{itemize}

Of course, a complementary measurement function is conceivable to revert the 
latter effect, but the intuition behind such a measurement needs some 
elaboration. We shall not discuss this issue in this paper.\\

Let us  also  define  the  following  functions  referred  to  as 
labelled Belief, labelled Plausibility and labelled Mass 
Functions respectively for the labeled population P:

\begin{df}
Let P be a population and $l$ its labeling. Then 

$$Bel_P    ^{M_l}(A)=\Prob{\omega}{P} \lnot M_l(\omega,\Xi-A)$$

$$Pl_P ^{M_l}(A)=\Prob{\omega}{P} M_l(\omega,A)$$

$$m_P ^{M_l}(A)=\Prob{\omega}{P} (\bigwedge_{B;B=\{v_i\}\subseteq A}
 M_l(\omega,B)
  \land \bigwedge_{B;B=\{v_i\}\subseteq \Xi-A} \lnot
 M_l(\omega,B))$$
\end{df}

Let us illustrate the above concepts with Citizen Coot example:

\begin{Bsp} \label{labelEx}
For the belief function for sold-bottles-population P  and the measurement 
function $M ^3$, let us assume the following labeling:\\
$l(\omega)=$\{(H,G),(H,B),(M,G),(M,B),(S,B),(D,B)\}\\
 for every $\omega \in \Omega$, which means that we are convinced that only 
high and moderate quality products are sold in good shops.
For the population P' under this labeling,
if we identify probability with relative frequency, we have 
the focal points given in the Table \ref{labtable}:
\end{Bsp}

\begin{table} \label{labtable} 
\caption{Mass and Belief Function under 
Modified
Measurement Method $M_{l} ^3$ }
\begin{center}
\begin{tabular}{|l|r|r|}
\hline
Set                           & $m_{P'} ^{M_{l} ^3}$ 
                              &  $Bel_{P'} ^{M_{l} ^3}$ \\
\hline
\{(H,B)                    \} &  20/717     &   20/717       \\
\{(H,G)                    \} & 112/717     &  112/717       \\
\{(H,B),(H,G)              \} &  70/717     &  202/717       \\
\{(M,B)                    \} &  80/717     &   80/717       \\
\{(M,G)                    \} & 127/717     &  127/717       \\
\{(M,B),(M,G)              \} & 110/717     &  317/717       \\
\{(S,B)                    \} &  65/717     &   65/717       \\
\{(D,B)                    \} &  13/717     &   13/717       \\
\{(H,B),(S,B)              \} &  15/717     &  100/717       \\
\{(H,B),(S,B),(H,G)        \} &  14/717     &  184/717       \\
\{(M,B),(S,B)              \} &  30/717     &  175/717       \\
\{(M,B),(S,B),(M,G)        \} &  25/717     &  387/717       \\
\{(H,B),(D,B)              \} &   8/717     &   41/717       \\
\{(H,B),(D,B),(H,G)        \} &   3/717     &  114/717       \\
\{(M,B),(D,B)              \} &  15/717     &  108/717       \\
\{(M,B),(D,B),(M,G)        \} &  10/717     &  228/717       \\
\hline
\end{tabular}
\end{center}
\end{table}

It is easily seen that:

\begin{th} 
$m_P ^{M_l}$ is the mass Function in the sense of 
DS-Theory.
\end{th}
\AnfBeweis To show this is suffices to show that the modified measurement 
method $M_l$ possesses the same properties as the measurement method $M$.\\
Let us consider a labeling $l$ and a population P under this labeling.\\
Let O be an object and L its label under labeling $l$ ($L=l(O)$).
Always  $M_l(O,\Xi)=TRUE$ because by definition $M_l(O,\Xi)=
M(O,\Xi \cap L)=M(O,L)$ and by definition of a labeled population for the 
object's O label L $M(O,L)=TRUE$.\\
Second, the superset consistency is satisfied, because if $A \subset B$ then
if  $M_l(O,A)=TRUE$ then also $M_l(O,A)=
M(O,A \cap L)=TRUE$, but because $A \cap L \subseteq B \cap L$ then also
$M(O,B \cap L)=TRUE$, but by definition $M(O,B \cap L)=M_l(O,B)=TRUE$ and 
thus it was shown that  $M_l(O,A)=TRUE$ implies  $M_l(O,B)=TRUE$ for 
any superset B of the set A.\\
Finally, also the subset consistency holds, because if 
 $M(O,L \cap A)=TRUE$ then there exists a proper subset B of $L \cap A$ such 
that  $M(O,B)=TRUE$. But in this case $B = L \cap B$ so we can formally write:
  $M(O,L \cap B)=TRUE$. Hence we see that $M_l(O,A)=TRUE$ implies the 
existence of a proper subset B of the set A such that $M_l(O,B)=TRUE$.
Hence considering analogies between definitions of $m_P ^M$ and $m_P {^M_l}$ 
as well 
as between the respective Theorems we see immediately that this Theorem  is 
valid.\\
\EndBeweis

\begin{th} 
$Bel_P ^{M_l}$ is a Belief Function in the sense of 
DS-Theory   corresponding  to the  $m_P ^{M_l}$. 
\end{th}
\AnfBeweis
As $M_l$ is shown to be a DS Theory Mass Function and considering 
 analogies between 
definitions of  $Bel_P ^M$ and $Bel_P {^M_l}$
as well as 
between the respective Theorems we see immediately that this Theorem  is 
valid.
\EndBeweis

\begin{th} 
 $Pl_P ^{M_l}$ is a Plausibility Function in the sense of 
DS-Theory and it is the Plausibility  Function  corresponding  to 
the  $Bel_P ^{M_l}$. 
\end{th}
\AnfBeweis
As $M_l$ is shown to be a DS Theory Mass Function and considering 
 analogies between 
definitions of  $Pl_P ^M$ and $Pl_P {^M_l}$
as well as 
between the respective Theorems we see immediately that this Theorem  is 
valid.
\EndBeweis

This does not complete the interpretation.  

Let  us  now  assume  we  run  a  "(re-)labelling  process"   on   the 
(pre-labelled or unlabeled)
population P. 

\begin{df}
Let $M$ be a measurement method, $l$ be a labeling under this measurement
method, and P be a population under this labeling (Note that the population
may also be unlabeled).
The  {\em (simple) labelling  process}   on    
the
population P 
is defined as a functional 
$LP: 2^\Xi \times \Gamma \rightarrow \Gamma$, where $\Gamma$ is the set of  
all  possible labelings under $M$, 
such that for the given labeling $l$ and a given nonempty
set of attribute values $L$ ($L  \subseteq \Xi$), 
it delivers a new labeling $l'$ ($l'=LP(L,l)$) such that for every object
$\omega \in \Omega$: 

1. if  $M_l(\omega,L)=FALSE$ then  
$l'(\omega)=\emptyset$\\
(that is l' discards a
labeled 
 object $(\omega,l(\omega))$ if $M_l(\omega,L )=FALSE$ 

2. otherwise $l'(\omega)=l(\omega) \cap L $
(that is l' labels the object with $l(\omega) \cap L $ otherwise.
\end{df}

Remark: It is immediately obvious, that the population obtained as the sample 
fulfills the requirements of the definition of a labeled population.\\

The labeling process clearly induces from P another population P' (a 
population under the labeling $l'$) being a subset of P (hence perhaps 
"smaller" 
than P)   labelled  a 
bit differently. Clearly if we  retain  the  primary  measurement 
method M then a  new  modified  measurement  method 
$M_{l'}$ is induced by the new labeling. 
The (re-)labelling process  may  be  imagined  as  the  diagnosis 
process made by a physician. A patient "labelled"  with  symptoms 
observed  by  himself  (many  symptoms  remain  hidden  for   the 
physician, like the body temperature curve over last few days) is 
relabeled by the physician when being ill (labelled with the diseases 
suspected)  or  rejected  (declared  healthy  due to  symptoms   not 
matching physician's diagnostic procedure).

Let us define  the  following 

\begin{df} "labelling  process  function" 
$m ^{LP;L }: 2 ^\Xi \rightarrow [0,1]$:
 is defined as:
 $$m ^{LP;L }(L )=1$$  
$$\forall_{B;  B  \in  2^\Xi,B \ne L } m ^{LP;L }(B)=0$$
\end{df}

It is immediately obvious that:

\begin{th} 
 $m ^{LP;L }$ is a Mass Function in sense of DS-Theory.
\end{th}

Let  $Bel  ^{LP,L }$  be  the  belief  and  $Pl  ^{LP,L }$  be  the 
Plausibility corresponding to $m ^{LP,L }$. Now let  us  pose  the 
question: what is the relationship between $Bel_{P'} ^{M_{l'}}$, 
 $Bel_P ^{M_l}$,  and $Bel ^{LP,L }$. It is easy to show that 

\begin{th} 
\label{thSimpleLab}
Let $M$ be a measurement function, $l$ a labeling, P a population under
this labeling. Let $L $ be a subset of $\Xi$. 
Let $LP$ be a labeling process and let $l'=LP(L ,l)$.
Let P' be a population under the labeling $l'$.
Then 
 $Bel_{P'} ^{M_{l'}}$ is a  combination  via  DS  Combination 
rule of  $Bel ^{M_l}$,  and $Bel ^{LP;L }$., that is:
$$Bel_{P'} ^{M_{l'}} = Bel_P ^{M_l} \oplus Bel ^{LP;L }$$.
\end{th}
\AnfBeweis
Let us consider a labeled object $(O_j,L_j)$ from the population P (before 
re-labeling, that is $L_j=l(O_j)$) which passed the relabeling and became
$(O_j,L_j \cap L )$, that is $L_j \cap L = l'(O_j)$..
Let us define $expr_B$ (before relabeling) and $expr_A$ (after labeling)
as:
$$expr_B((O_j,L_j),A) =  \bigwedge_{B;B=\{v_i\}\subseteq A} M_l(O,B)
  \land$$
$$ \land
 \bigwedge_{B;B=\{v_i\}\subseteq \Xi-A} \lnot  M_l(O,B)$$
and 
$$expr_A((O_j,L_j),A) =  \bigwedge_{B;B=\{v_i\}\subseteq A} M_{l'}(O
 ,B) \land$$
$$  \land \bigwedge_{B;B=\{v_i\}\subseteq \Xi-A} \lnot  M_{l'}(O ,B)$$
Let $expr_B((O_j,L_j),C)=TRUE$ and $expr_A((O_j,L_j),D)=TRUE$ for some C and 
some D. Obviously then for no other C and no other D the respective 
expressions are valid. It holds also that:
$$expr_B((O_j,L_j),C) =  \bigwedge_{B;B=\{v_i\}\subseteq C} M(O_j,L_j \cap B)
\land $$
$$  \land \bigwedge_{B;B=\{v_i\}\subseteq \Xi-D} \lnot  M((O_j,L_j \cap B)$$
and 
$$expr_A((O_j,L_j),D) =  \bigwedge_{B;B=\{v_i\}\subseteq D} M(O_j,L_j \cap 
L 
 \cap B)
  \land$$
$$\land
 \bigwedge_{B;B=\{v_i\}\subseteq \Xi-D} \lnot  M(O_j,L_j \cap 
L  \cap B)$$
In order to get truth on the first expression, C must be a subset of $L_j$, 
and for the second we need D to be a subset of $L_j \cap L $.
Furthermore, for a singleton $F \subseteq \Xi$ 
either $M(O_j,L_j \cap F)=TRUE, M(O_j,L_j \cap L   \cap F)=TRUE$, and then
it belongs to C, $L  $ and D, or
  $M(O_j,L_j \cap F)=TRUE, M(O_j,L_j \cap L   \cap F)=FALSE$, and then
it belongs to C, but not to $L  $ and hence not to D, or
  $M(O_j,L_j \cap F)=FALSE$, so due to superset consistency also
 $M(O_j,L_j \cap L   \cap F)=FALSE$, and then
it belongs  neither to C nor to D
(though membership in $L $ does not need to be excluded).
So we can state that  $D = C \cap L $,

So the absolute expected frequency of objects for which $expr_A(D)$ holds, is 
given by:
$$ \sum_{C; D= C \cap L } samplecardinality \cdot  m_P ^{M_l}(C)$$
that is:\\
$$ \sum_{C; D= C \cap L } samplecardinality \cdot  m_P ^{M_l}(C)\cdot 
m ^{LP;L }(L)$$
which can be easily re-expressed as: \\
$$ \sum_{C,G; D= C \cap G} samplecardinality \cdot  m_P ^{M_l}(C)\cdot 
m ^{LP;L }(G)$$
So generally:\\
$$ m_{P'}^{M_{l'}}(D) =c\cdot  \sum_{C,G; D= C \cap G}
  m_P ^{M_l}(C)\cdot m ^{LP;L }(G)$$
with c - normalizing constant.
\EndBeweis

\begin{Bsp} To continue Citizen Coot example let us recall
the function $Bel_P ^{M}$ from Example \ref{nonlabelEx} which is one of 
an unlabeled population.
Let us define the label $$L=
\{(H,G),(H,B),(M,G),(M,B),(S,B),(D,B)\}$$
 as in Example \ref{labelEx}. Let us define the labeling process function as 
 $$m ^{LP;L }(L )=1$$  
$$\forall_{B;  B  \in  2^\Xi,B \ne L } m ^{LP;L }(B)=0$$.
Let us consider the function  $Bel_{P'} ^{M_l}$ from Example \ref{labelEx}.
It is easily seen that:
$$Bel_{P'} ^{M_l}=Bel_{P}^{M} \oplus  Bel ^{LP;L }$$
%
\end{Bsp}

Let us try  another  experiment, with a more general (re-)labeling process. 
  Instead  of  a  single  set  of 
attribute  values  let  us  take  a  set  of  sets  of  attribute 
values $L ^1, L ^2, ...,L ^k$  (not  necessarily  disjoint)  and 
assign to each one a probability 
$m ^{LP, L ^1, L ^2, ...,L ^k}(A_i)$
of selection.

\begin{df}
Let $M$ be a measurement method, $l$ be a labeling under this measurement
method, and P be a population under this labeling (Note that the population
may also be unlabeled).
Let  us  take  a  set  of (not  necessarily  disjoint) nonempty sets  of  
attribute values $\{L ^1, L ^2, ...,L ^k\}$    and 
let us define the  probability of selection as a function
$m ^{LP, L ^1, L ^2, ...,L ^k}: 2 ^\Xi \rightarrow [0,1]$ such that
$$\sum_{A;A \subseteq \Xi}m ^{LP, L ^1, L ^2, ...,L ^k}(A)=1$$
$$\forall_{A; A \in \{ L ^1, L ^2, ...,L ^k\}} 
m ^{LP, L ^1, L ^2, ...,L ^k}(A)>0$$
$$\forall_{A; A \not\in \{ L ^1, L ^2, ...,L ^k\}} 
m ^{LP, L ^1, L ^2, ...,L ^k}(A)=0$$
 The  {\em (general) labelling  process}   on    
the
population P 
is defined as a (randomized) functional 
$LP: 2^{2^\Xi} \times \Delta
\times  \Gamma \rightarrow \Gamma$, where $\Gamma$ is the set 
of all  possible labelings under $M$, and $\Delta$ is 
a set of all possible probability of selection functions,
such that for the given labeling $l$ and a given 
 set  of (not  necessarily  disjoint) nonempty sets  of  
attribute values $\{L ^1, L ^2, ...,L ^k\}$    and 
a given probability of selection 
$m ^{LP, L ^1, L ^2, ...,L ^k}$
it delivers a new labeling $l"$ such that for every object
$\omega \in \Omega$:

1. a label L, element of the set $\{ L ^1, L ^2, ...,L ^k\}$ 
is sampled randomly according to the probability distribution 
$m ^{LP, L ^1, L ^2, ...,L ^k}$;
This sampling is done independently for each individual object,

2. if  $M_l(\omega,L)=FALSE$ then  
$l"(\omega)=\emptyset$\\
(that is l" discards an object $(\omega,l(\omega))$ if 
$M_l(\omega,L )=FALSE$ 

3. otherwise $l"(\omega)=l(\omega) \cap L $
(that is l" labels the object with $l(\omega) \cap L $ otherwise.)
\end{df}

Again we obtain another ("smaller") population P" under the labeling $l"$  
labelled 
 a bit differently. Also a  new  modified  measurement  method 
$M_{l"}$ is induced by the "re-labelled" population.  
Please notice, that $l"$ is not derived deterministicly. 
Another run of the general (re-)labeling process LP may result in a different
final labeling of the population and hence a different subpopulation under 
this new labeling.

\begin{Bsp}
The (re-)labelling process  may  be  imagined  as  the  diagnosis 
process made by a team of physicians in a poly-clinic. A patient "labelled" 
with  symptoms 
observed  by  himself 
is (a bit randomly) directed by the ward administration to one of 
the available internists each of  them  having  a  bit  different 
educational background  and/or  a  different  experience  in  his 
profession, hence taking into consideration a bit  different  set 
of working hypotheses. The patient is 
relabeled by the given physician being  ill  (labelled  with  the 
diseases 
suspected)  or  rejected  (declared  healthy)  according  to   the 
knowledge of this particular physician. 
The final ward statistics of illnesses does not take
into account the fact that a physician may have had no knowledge
of a particular disease unit and hence qualified the patient
either healthy or ill of another, related disease unit.
And it reflects the combined processes: of  random allocations
of patients to physicians and of belief worlds of the physicians
rather then what the patients were actually suffering from.
(We are actually satisfied with the fact that both views of 
ward statistics usually converge).
\end{Bsp}

Clearly:

\begin{th} 
$m ^{LP,L ^1,...,L ^k}$ is a Mass Function in sense of DS-Theory.
\end{th}

Let   $Bel   ^{LP;L ^1,...,L ^k}$   be   the   belief    and    $Pl 
^{LP,L ^1,...,L ^k}$  be  the 
Plausibility corresponding to $m ^{LP,L ^1,...,L ^k}$. Now let  us  pose  the 
question: what is the relationship between $Bel_{P"} ^{M_{l"}}$, 
 $Bel_P ^{M_l}$,  and $Bel ^{LP,L ^1,...,L ^k}$. It is easy to show that 

\begin{th} 
Let $M$ be a measurement function, $l$ a labeling, P a population under
this labeling. 
Let $LP$ be a generalized labeling process and let $l"$
be the result of application of the $LP$ for the set
of labels from the set $\{ L ^1, L ^2, ...,L ^k\}$ 
 sampled randomly according to the probability distribution 
$m ^{LP, L ^1, L ^2, ...,L ^k}$;.
Let P" be a population under the labeling $l"$.
Then 
The expected value 
over the set of all possible resultant labelings $l"$ (and hence
populations P") 
(or, more precisely, value vector) of 
$Bel_{P"} ^{M_{l"}}$ is a  combination  via  DS  Combination 
rule of  $Bel_P ^{M_l}$,  and $Bel ^{LP,L ^1,...,L ^k}$., that is:
$$E(Bel_{P"} ^{M_l'}) = Bel_P ^{M_l} \oplus Bel ^{LP,L ^1,...,L ^k}$$.
\end{th}
\AnfBeweis
By the same reasoning as in the  proof 
of Theorem \ref{thSimpleLab} 
we come to the 
conclusion that for the given label  $L ^i$ and the labeling $l"$ (instead of 
$l'$ 
 the absolute expected frequency of objects for which $expr_A(D)$ holds, is 
given by:
$$ \sum_{C; D= C \cap L ^i} samplecardinality \cdot  m_P ^{M_l}(C)
  \cdot  m ^{LP;L ^1,...,L ^k}(L ^i)
$$
as the process of sampling the population runs independently of the sampling 
the set of labels of the labeling process.\\
But  $expr_A(D)$ may hold for any   $L ^i$ such that   $C \subseteq L ^i$, 
hence in all the   $expr_A(D)$ holds for as many objects as:\\
$$ \sum_{i;i=1,...,k} \sum_{C; D= C \cap L ^i} samplecardinality \cdot  m_P 
^{M_l}(C) \cdot m ^{LP;L ^1,...,L ^k}(L ^i)
$$

which can be easily re-expressed as: \\
$$ \sum_{C,G; D= C \cap G} samplecardinality \cdot  m_P ^{M_l}(C)\cdot 
m ^{LP;L ^1,...,L ^k}(G)$$
So generally:\\
$$ E(m_{P"}^{M_{l"}}(D) )=c\cdot  \sum_{C; D= C \cap G}  m_P ^{M_l}(C)\cdot 
m ^{LP;L ^1,...,L ^k}(G)$$
with c - normalizing constant.
Hence the claimed relationship really holds.
 \EndBeweis

\begin{Bsp}
The generalized labeling process and its consequences
may be realized in our Citizen Coot example 
by randomly assigning the sold bottles for evaluation to two
"experts", one of them 
- considering about 30 \% of the bottles - is running the full $M$ test 
procedure, and  the other - having to  consider the remaining 70 \% of 
 checked bottles - makes it easier for himself  by making use of his belief 
in the labeling $l$ of Example \ref{labelEx}.  
\end{Bsp}

%

\subsection{Summary of the New Interpretation}

     The following results have been established in this Section:
\begin{itemize}
\item  concepts of measurement and modified measurement methods  have 
been introduced
\item a concept of labelled population has been developed
\item  it has been shown that a labelled population with the modified 
measurement method can be considered in terms of a  Joint  Belief 
Distribution in the sense of DS-Theory,
\item  the process of "relabeling" of a labelled population has been 
defined and shown to be describable as a Belief Distribution.
\item  it has been shown that the  relationship  between  the  Belief 
Distributions of the resulting relabeled population,  the  basic 
population and the relabeling process can be expressed in terms 
of the Dempster-Rule-of-Independent-Evidence-Combination. 
\end{itemize}

     This  last  result  can  be  considered  as  of   particular 
practical importance. The interpretation schemata of DS  Theory  made  by 
other authors suffered from one basic shortcoming:
if we interpreted population data as well as evidence in terms of 
their DS schemes, and then combine the evidence  with  population 
data (understood as a Dempster type  of  conditioning)  then  the 
resulting belief function cannot be interpreted in terms  of  the 
population data scheme,  with  subsequent  updating  of  evidence 
making thinks worse till even the weakest  relation  between  the 
belief function and the (selected sub)population is lost.

In this paper we achieve a  break-through:  data  have  the  same 
interpretation scheme after any number of evidential updating and 
hence the belief function can be verified against the data at any 
moment of DS evidential reasoning.

The above definition and properties of the generalized labeling process  
should 
be considered from a philosophical point of view. If we take one by one the 
objects of our domain, possibly labelled previously by an expert in the past,
 and assign a label independently of the actual value of 
the attribute of the object, then we cannot claim in any way that such a 
process may be attributed to the opinion of the expert. Opinions of two 
experts may be independent of one another, but they cannot be independent
of the subject under consideration. This is the point of view with which most 
people would agree, and should the opinions of the experts not depend
on the subject, then at least one of them may be considered as not expert.

This is exactly what we want to point at with our interpretation: 
the precise pinpointing at what kind of independence is assumed 
within the Dempster-Shafer theory is essential for its usability.
Under our interpretation, the independence relies in trying to 
select  a label for fitting to 
an object independently of whatever properties this object has (including its 
previous labeling). The distribution of labels for fitting is exactly 
identical from object to object. The point, where the dependence of object's 
labeling on its properties comes to appearance, is  when the measurement 
method 
states that the label does not fit. Then the object is discarded. From 
philosophical point of view it means exactly that we try to impose our 
 philosophy of life onto the facts: cumbersome facts are neglected and 
ignored. We suspect that this is exactly the justification of the name "belief 
function". It expresses not what we see but what we would like to see.\\
Our suspicion is strongly supported by the quite recent statement of Smets 
that "authors (of multiple interpretations in terms of
upper lower probability models, inner and outer measures, 
random sets, probabilities of provability, 
probabilities of necessity etc.)
usually do not explain or justify the dynamic component, that  is, how 
updating (conditioning) is to be handled (except in 
some cases by defining conditioning as a special case of combination. So I 
 (that is Smets) feel that these partial comparisons are incomplete, 
especially as all these interpretations lead to different updating rules. " 
Our interpretation explains both the static and dynamic component of the DST, 
and does not lead to any other but to the Dempster Rule of Combination, hence 
 may be acceptable from the rigorous point of view of Smets. As in the light 
of Smets' paper \cite{Smets:92} we  have presented the only correct 
probabilistic interpretation of the DS theory so far, we feel to be 
authorized to claim that our philosophical assessment of the DST is 
the correct one. \\

We have seen from the proofs of the theorems of this paper, that our 
interpretation may be called {\em a} true one. The paper of Smets 
\cite{Smets:92} permits us to claim that we have found {\em the} true 
interpretation.

\section{Belief from Data}

As the DS-belief function introduced in this paper is defined in terms of 
frequentist measures, there exists a direct possibility of calculating the 
belief function from data. \\

It has to be assumed that we have a data set for which the measurements of 
type $M_l$ have been carried out for each singleton subset of the space of 
discourse $\Xi$. The results of these measurements may be available for 
example as a set-valued attribute associated with each object in such a 
way 
that the  values actually appearing are those for which the singleton set  
tests were positive 
(i.e. TRUE). In this case if for an object the attribute $X$ has the value
$X=A$ with $A \subseteq \Xi$ then this object increases the count for the 
DS-Mass Function $m(A)$ (and for no other m).\\


Whenever any statistical quantity is estimated from data, there exists some 
risk (uncertainty) about unseen examples. If we assume some significance 
levels, we can complete the estimation by taking the lower bounds as actual 
estimates of m's and shifting the remaining burden (summing up to 1) onto the 
$m(\Xi)$ just taking for granted that doubtful cases may be considered as 
matching all the measurements.

\section{Discussion}
In the past, various interpretations have been sought for the 
Dempster-Shafer
Bel-Functions. Two main steams of research were distinguished by Smets 
\cite{Smets:92}: probability related approaches and probability discarding 
approaches (the former disguised, the latter welcome by Smets). Let us make 
some comparisons with our interpretation and its underlying philosophy.
 
\subsection{Shafer and Smets}

Shafer \cite{Shafer:90b} and Smets \cite{Smets:92} have made some strong 
statements in defense of the Dempster-Shafer theory against sharp criticism 
of this theory by its opponents as well as unfortunate users of the DST who 
wanted to attach it to the dirty reality (that is objectively given 
databases). Smets \cite{Smets:92} and also initially Shafer \cite{Shafer:76} 
insisted on Bels  not being connected to any empirical measure (frequency, 
probability 
etc.) considering the domain of DST applications  as the one where 
"we are ignorant of the existence of probabilities", and not one with 
         "poorly known probabilities" (\cite{Smets:92}, p.324). 
The basic property of probability, which should be dropped in the DST 
axiomatization, should be  the additivity of belief measures.  
Surely, it is easily possible to imagine situations where in the real life 
additivity is not granted: 
imagine we have had a cage with 3 pigs, we put into it 3 
hungry lions two hours ago, how many animals are there now ? ($3+3 <6$). Or 
ten years ago we left one young man and one young woman on an island in the 
 middle of the atlantic ocean with food and weapons sufficing for 20 years. 
How many human beings are there now ? ($1+1>2$). \\
The trouble is, however, that the objects stored in databases of a computer 
behave usually (under normal operation) in an additive manner. Hence the DST 
is simply disqualified for any reasoning within human collected data on real 
world, if we accept the philosophy of Smets and Shafer. 

The question may be raised at this point, what else practically useful 
can be obtained 
from a computer reasoning on the basis of such a DST. 
If the DST models, as Smets and Shafer claim, human behaviour during 
evidential reasoning, then  
it would have to be 
demonstrated that humans indeed reason as DST. 
We take e.g. 1000 people who never heard of Dempster-Shafer theory,
briefly explain the static component, provide them with two opinions of 
 independent experts and expect of them to answers what are their final 
beliefs.
Should their answers correspond to results of the DST (at least converge 
toward them), then the computer, if fed 
with our knowledge, would 
be capable to predict our conclusions on a given subject. However, to my 
knowledge, no experiment like this has ever been carried out. 
 Under these circumstances 
the computer reasoning with DST would tell us what we have to think and 
not 
what we think. But I don't suspect that anybody would be happy about a 
computer like this.

Hence, from the point of view of computer implementation the 
philosophy of Smets and Shafer is not acceptable.
Compare also Discussion in \cite{Halpern:92} on the subject.

Both of them felt a bit uneasy about a total loss of reference to any 
scientific experiment checking practical applicability of the DST and 
suggested some probabilistic background for decision making (e.g. the 
pigeonistic probabilities of Smets), but I am afraid that by these 
interpretations they fall precisely into the same pitfalls they claimed to 
avoid by their highly abstract philosophy.

As statistical properties of Shafer's \cite{Shafer:76} notion of evidence are 
concerned, sufficient criticism has been expressed by Halpern and Fagin 
(\cite{Halpern:92} in sections 4-5). Essentially it is
pointed there at the fact that "the belief that represents the joint 
observation is equal to the combination is in general not equal to the 
combination of the belief functions representing the individual (independent) 
observations" (p.297). The other point raised there that though it is possible 
to capture properly in belief functions evidence in terms of 
probability  of observations update functions (section 4 of 
\cite{Halpern:92}), it is not possible to do the same if we would like to 
capture evidence in terms of beliefs  of observations update functions  
(section 5 of \cite{Halpern:92}). 

As Smets probabilistic interpretations are concerned, let us "continue" the 
killer example of \cite{Smets:92} on pages 330-331. "There are three 
potential killers, A, B, C. Each can use a gun or a knife. I shall select one 
of them, but you will not know how I select the killer. The killer selects 
 his weapon by a random process with p(gun)=0.2 and p(knife)=0.8. Each of A, 
B, C 
has his own personal random device, the random devices are unrelated. ...... 
Suppose you are a Bayesian and you must express your "belief" that the killer 
will use a gun. The BF (belief function) solution gives $Bel(gun)=0.2 \times 
0.2 \times 0.2=0.008$. ..... Would you defend 0.2 ? But this applies only if I 
select a killer with a random device ...... But I never said I would use a 
random device; I might be a very hostile player and cheat whenever I can. ... 
. So you could interpret bel(x) as the probability that you are sure to win 
whatever Mother Nature (however hostile) will do." \\
Yes, I will try to continue the hostile Mother Nature game here. For 
completeness I understand that $Bel(knife)=0.8 ^3=0.512$ and 
$Bel(\{gun,knife\})=1$. But suppose there is another I, the chief of gangster
science fiction 
physicians, making decisions independly of the chief I of the killers. The 
chief I of physicians knows of the planned murder and has three physicians 
X,Y,Z.  Each can either rescue a killed man or let him die. 
I shall select one 
of them, but you will not know how I select the physician. The physician, in 
case of killing with a gun, selects 
his attritude by a random process with $p(rescue|gun)=0.2$ 
and $p(let\  die|gun)=0.8$ and he lets the person die otherwise. Each 
of X, Y, Z 
has his own personal random device, the random devices are unrelated. ...... 
Suppose you are a Bayesian and you must express your "belief" that the 
physician will rescue if the killer 
will use a gun. The BF (belief function) solution gives 
$Bel_1(rescue|gun)=0.2^3=0.008$. $Bel_1(let\  die|gun)=0.8^3=0.512$, 
$Bel_1(\{recue,let \  die\}|gun)=1$. Also  $Bel_2(let\  
die|knife)=1$. As the scenarios for $Bel_1$ and $Bel_2$ are independent, let 
us combine them by the Dempster rule: $Bel_{12}=Bel_1 \oplus Bel_2$. We make 
use of the Smets' claim that "the de re and de dicto interpretations lead to 
the same results" (\cite{Smets:92}, p. 333), that is $Bel(A|B)=Bel(\lnot B 
\lor A)$. Hence 
$$m_{12}(\{(gun,let\ die),(knife,let\ die),(knife,rescue)\})=0.480$$
$$m_{12}(\{(gun,rescue),(knife,rescue)\})=0.008$$
$$m_{12}(\{(knife,rescue),(gun,let\ die)\})=0.512$$

Now let us combine $Bel_{12}$ with the original $Bel$. We obtain:\\
$$m\oplus m_{12}((gun,let\ die)=0.008 \cdot 0.480+0.008 \cdot 0.512=
0.008 \cdot 0.992$$

But these two unfriendly chiefs of gangster organizations can be extremely 
unfriendly and in fact your chance of winning a bet may be as bad as $0.008 
\cdot 0.512$ for the event $(gun,let\ die)$. Hence the "model" proposed by 
Smets for understanding beliefs functions as "unfriendly Mother Nature" is 
simply wrong. If the Reader finds the combination of $Bel_2$ with the other 
Bels a little tricky, then for justification He should refer to the paper of 
Smets and have a closer look at all the other examples. \\

Now returning to the philosophy of "subjectivity" of Bel measures:
Even if a human being may 
possess his private view on a subject, it is only after we formalize the 
feeling of subjectiveness and hence ground it in the data that we can rely on 
 any computer's "opinion". We hope we have found one such formalization in 
this 
paper. The notion of labeling developed here substitutes one aspect of 
subjective human behaviour - if one has found one plausible explanation, one 
is too lazy to look for another one. So the process of labeling may express 
our personal attitudes, prejudices, sympathies etc. The interpretation drops 
deliberately the strive for maximal objectiveness aimed at by traditional 
statistical analysis. Hence we think this may be a promising path for further 
research going beyond the DS-Theory formalism.

Smets \cite{Smets:92} views  the probability theory as a formal mathematical 
apparatus and hence puts it on the same footing as his view of the DST. 
However, in our opinion, he ignores totally one important thing: The abstract 
concept of probability has its real world counterpart of relative frequency 
which tends to behave approximately like the theoretical probability in 
sufficiently many experimental settings as to make the abstract concept of 
probability useful for practical life. And a man-in-the-street will expect of 
the DST to possess also such a counterpart or otherwise the DST will be 
 considered as another version of the theory of counting devils on a 
pin-head.\\ 

Let us also have a look at interpretations disguised by Shafer and Smets 
(i.e. all the mentioned below):\\

\subsection{DST and Random Sets}

The canonic 
random set 
interpretation 
\cite{Nguyen:78}
is one with a statistical process over set
instantiations. The rule of combination assumes then that two such 
statistically independent processes are run and we are interested in their 
intersections. This approach is not sound as empty intersection is excluded  
and this will render any two processes statistically dependent.
We overcome this difficulty assuming in a straight forward manner that 
we are "walking" from population to population applying the Rule of 
Combination. Classical DS theory in fact assumes such a walk implicitly
or it drops in fact the assumption that Bel() of the empty set is equal 0.
In this sense the random set approaches may be considered as sound as ours. 

However, in many cases the applications of the model are insane. 
 For example, 
to imitate the logical inference it is frequently assumed that we have a 
Bel-function describing the actual observed value of a predicate P(x), 
 and a Bel-Function describing the implication "If P(x) then Q(x)"
\cite{Ma:91}. It is 
assumed further that the evidence on the validity of both Bel's has been 
collected independently and one applies the DS-rule of combination to 
calculate the Bel of the predicate Q(x). One has then to assume that there is 
a focal m of the following expression: $m(\{(P(x) , Q(x)),
(\lnot P(x) , Q(x)),(\lnot P(x) ,\lnot  Q(x)) \})$ which actually means that 
with non-zero probability at the same time $P(x)$ and $\lnot P(x)$ hold for 
 the same object as we will see in the following example:  
Let $Bel_1$ represent our belief in the implication, with focal points:
$$m_1(P(x) \rightarrow Q(x))=0.5, 
\ m_1(\lnot (P(x) \rightarrow Q(x)))=0.5, $$
Let further the independent opinion $Bel_2$ on P(x) be available in the form 
of focal points:
$$m_2(P(x))=0.5, \ m_2(\lnot P(x))=0.5$$
Let $Bel_{12}=Bel_1 \oplus Bel_2$  represent the combined opinions of both 
experts. The focal points of $Bel_{12}$ are:
$$ m_{12}(\{(P(x) , Q(x))\})=0.33, \ 
   m_{12}(\{(P(x) ,\lnot Q(x))\})=0.33, $$  
$$   m_{12}(\{(\lnot P(x) , Q(x)),(\lnot P(x) ,\lnot  Q(x)) \})=0.33$$ 

$m_{12}(\{(P(x) , Q(x))\})=0.33$  makes us believe that there exist objects 
for 
which both P(x) and Q(x) holds. However, a sober (statistical) look at expert 
opinions suggests that  all situations for which the implication $P(x) 
\rightarrow Q(x)$ holds, must result from falsity of $P(x)$, hence whenever 
Q(x) holds then $\lnot P(x)$ holds. These two facts combined mean that P(x) 
and its negation have to hold simultaneously. 
 This is actually absurdity overseen deliberately. The source of this 
 misunderstanding is obvious: the lack of proper definition of what is and 
what is  not independent. 
 Our interpretation allows 
for 
sanitation of this situation. We are not telling that the predicate and its 
negation hold simultaneously. Instead we say that
for one object we modify 
 the measurement procedure (set a label) in such a way that it,
applied for calculation of $P(x)$, yields true and at the same time 
for another object, with the same original properties 
we make another modification of 
measurement procedure (attach a label to it) so that 
 measurement of $\lnot P(x)$ yields also 
true, because possibly two different persons were enforcing their different 
beliefs onto different subsets of data.\\

Our approach is also superior to canonical random set approach in the 
following sense: The 
canonical approach requires knowledge of the complete  random  set 
realizations 
of two processes on an object to determine the combination of both processes. 
 We, however, postpone the acquisition of
 knowledge of the precise 
instantiation of properties 
of the object by interleaving the concept of measurement and the concept of 
labeling process. This has a close resemblance to practical processing 
whenever diagnosis for a patient is made. If a physician finds a set of 
hypotheses explaining the symptoms of a patient, he will 
usually not
 try to carry 
out other testing procedures than those related to the plausible hypotheses.
He runs clearly at risk that there exists a different set of hypotheses
also explaining the patients's symptoms, and so a disease unit possibly 
present may not be detected on time, but usually the risk is sufficiently
low to proceed in this way, and the cost savings may prove enormous. \\

\subsection{Upper and Lower Probabilities}

Still another approach was to handle Bel and Pl as lower and upper 
probabilities
 \cite{Dempster:67}. This approach is of limited use as not every set of 
lower and upper probabilities leads to Bel/Pl functions \cite{Kyburg:87},
hence establishing a unidirectional relationship between probability theory 
and the DS-theory. Under our interpretation, 
the Bel/Pl function pair may be considered as a kind of interval 
approximations to some "intrinsic" probability distributions which, however, 
cannot be accessed by feasible  measurements and are only of interest as a 
kind 
of qualitative explanation to the physical quantities really measured.\\

Therefore another approach was to handle them as lower/upper envelops to some 
probability density function realization  \cite{Kyburg:87},  
  \cite{Fagin:91B}. However, the DS rule of combination of independent 
evidence 
failed. \\

\subsection{Inner and Outer Measures}

Still another approach was to handle Bels/Pl in probabilistic structures 
rather than in probabilistic spaces \cite{Fagin:91}. Here, DS-rule could be 
justified as 
one of the possible outcomes of independent combinations, but no stronger 
properties were available. This is due to the previously mentioned fact that 
exclusion of empty intersections renders actually most of conceivable 
processes dependent. Please notice that under our interpretation no 
such ambiguity occurs. This is because we not only drop empty intersecting 
 objects but also relabel the remaining ones so that any probability 
calculated 
afterwards does not refer to the original population.

So it was tried to drop the DS-rule altogether in the probabilistic 
structures, but then it was not possible to find a meaningful rule for 
multistage  reasoning \cite{Halpern:92}. This is a very important negative 
outcome. As the Dempster-Shafer-Theory is sound in this respect and possesses 
many  useful properties (as mentioned in the Introduction), it should be 
sought 
for an interpretation meeting the axiomatic system of DS Theory rather then 
tried to violate its fundamentals. Hence we consider our interpretation as a 
promising one for which decomposition of the joint distribution paralleling 
the results for probability distributions may be found based on the data.\\

\subsection{Rough Set Approach}

An interesting alternative interpretation of the Dempster-Shafer Theory was 
found within the framework of the rough set theory \cite{Skowron:93}, 
\cite{Grzymala:91}. Essentially the rough set theory searches for 
approximation of the value of a decision attribute by some other
(explaining) attributes. 
 It usually happens that those attributes are capable only of providing a 
lower 
and upper approximation to the value of the decision attribute (that is the 
set of vectors of explaining attributes 
supporting only this value 
 of the decision variable,
and the set of vectors of explaining attributes 
supporting also this value 
 of the decision variable resp.- for details 
see texts of Skowron \cite{Skowron:93} and Grzyma{\l}a-Busse 
\cite{Grzymala:91}).
The Dempster Rule of combination is interpreted by Skowron \cite{Skowron:93b} 
as combination of opinions of independent experts, who possibly look at 
different sets of explanation attributes and hence may propose different 
explanations. 

The difference between  our approach and the one based on rough sets lies 
first of all in the ideological background: We assume that the "decision 
attribute" is set-valued whereas the rough-set approach assumes it to be 
single-valued. This could have been overcome by some tricks which will not be 
explained in detail here.
But the combination step is here essential: If 
we assume that the data sets for forming  knowledge of these two experts are 
exhaustive, then it can never occur that these opinions are contradictory. 
 But 
the DST rule of combination uses the normalization factor for dealing with 
cases like this.
Also the opinions of experts may have only the form of a simple (that is 
deterministic) support function. Hence, rough-set interpretation implies 
axioms not actually present in the DST. Hence rough set interpretation is 
 on the one hand restrictive, and on the other hand not fully conforming to 
the general DST. From our point of view the DST would change the values of 
decision variables rather then recover them from expert opinions.

Here, we come again at the problem of viewing the independence of experts.  
The DST assumes some strange kind of independence within the data: the 
proportionality of the distribution of masses of sets of values among 
intersecting subsets weight by their masses in the other expert opinion.  
Particularly unhappy is the fact for the rough set theory, that given a 
value of the decision variable, the respective indicating vectors of 
explaining variables values must be proportionally distributed among the 
experts not only for this decision attribute value, but also for all the 
other decision attribute values that ever belong to the same focal point.  
Hence applicability of the rough set approach is hard to justify by a 
simple(, "usual"  as Shafer wants) statistical test. 
On the other hand, statistical independence required for Dempster rule 
application within our approach is easily checked.

To demonstrate the problem of rough set theory with re combination of opinions 
of independent experts let us consider an examle of two experts having the 
combined explanatory attributes $E_1$ (for expert 1) and $E_2$ (for expert 2) 
both trying to guess the decision attribute $D$. Let us assume that $D$ takes 
one of two values: $d_1,d_2$, $E_1$ takes one of three values $e_{11}, e_{12}, 
e_{13}$, $E_2$ takes one of three values $e_{21}, e_{22}, 
e_{23}$. Furthermore let us assume that the rough set analysis
of an exhaustive set of possible cases
 shows that the 
value $e_{11}$ of the attribute $E_1$ indicates the value $d_1$ of the 
decision attribute $D$, 
$e_{12}$ indicates $d_2$,
$e_{13}$ indicates the set \{$d_1,d_2$\},
 Also        let us assume that the rough set analysis of an exhaustive set of 
possible cases shows that the 
value $e_{21}$ of the attribute $E_2$ indicates the value $d_1$ of the 
decision attribute $D$, 
$e_{22}$ indicates $d_2$,
$e_{32}$ indicates the set \{$d_1,d_2$\},
 From the point of view of 
bayesian analysis four cases of causal influence may be distinguished (arrows 
indicate the direction of dependence).
$$E_1 \rightarrow D \rightarrow  E_2$$
$$E_1 \leftarrow  D \leftarrow   E_2$$
$$E_1 \leftarrow  D \rightarrow  E_2$$
$$E_1 \rightarrow D \leftarrow   E_2$$

 From the point of view of bayesian analysis, in the last case attributes 
$E_1$ and $E_2$ have to be unconditionally independent, in the remaining 
cases: $E_1$ and $E_2$ have to be independent  conditioned on $D$. 
Let us consider first unconditional independence of $E_1$ and $E_2$. Then 
we have tthat:
$$(\Prob{\omega}{P} E_1(\omega)=e_{11} \land E_2(\omega)=e_{22}) = $$
$$=
(\Prob{\omega}{P} E_1(\omega)=e_{11} ) \cdot 
(\Prob{\omega}{P}  E_2(\omega)=e_{22}) >0  $$
 However, it is impossible that $(\Prob{\omega}{P} E_1(\omega)=e_{11} \land 
E_2(\omega)=e_{22}) > 0$  because we have to do with experts who may provide 
us possibly with information not specific enough, but will never provide us 
with 
contradictory information. We conclude that unconditional independence of 
experts is impossible.\\
Let us turn to independence of $E_1$ and $E_2$ if  conditioned on $D$. 
We introduce the following denotation:\\
$$p_1 = \Prob{\omega}{P} D(\omega)=d_1$$
$$p_2 = \Prob{\omega}{P} D(\omega)=d_2$$
$$e_1'= \Prob{\omega}{ (D(\omega)=d_1)\land P} E_1(\omega)=e_{11}$$
$$e_3'= \Prob{\omega}{ (D(\omega)=d_1)\land P} E_1(\omega)=e_{13}$$
$$f_1'= \Prob{\omega}{ (D(\omega)=d_1)\land P} E_2(\omega)=e_{21}$$
$$f_3'= \Prob{\omega}{ (D(\omega)=d_1)\land P} E_2(\omega)=e_{23}$$
$$e_2"= \Prob{\omega}{ (D(\omega)=d_2)\land P} E_1(\omega)=e_{12}$$
$$e_3"= \Prob{\omega}{ (D(\omega)=d_2)\land P} E_1(\omega)=e_{13}$$
$$f_2"= \Prob{\omega}{ (D(\omega)=d_2)\land P} E_2(\omega)=e_{22}$$
$$f_3"= \Prob{\omega}{ (D(\omega)=d_2)\land P} E_2(\omega)=e_{23}$$
 Let $Bel_1$ and $m_1$ be the belief function and the mass function 
representing the knowledge of the first expert, let $Bel_2$ and $m_2$ be the 
belief function and the mass function 
representing the knowledge of the second expert. Let $Bel_{12}$ and $m_{12}$ 
be the belief function and the mass function 
representing the knowledge contained in the combined usage of attributes 
$E_1,E_2$ if used for prediction of $D$ - on the grounds of the rough set 
theory. It can be easily checked that:\\
$$m_1(\{d_1\})=e_1' \cdot p_1,\  m_1(\{d_2\})=e_2" \cdot p_2,\ 
m_1(\{d_1,d_2\})=e_3' \cdot p_1, + e_3"' \cdot p_2$$
$$m_2(\{d_1\})=f_1' \cdot p_1,\  m_2(\{d_2\})=f_2" \cdot p_2,\ 
m_2(\{d_1,d_2\})=f_3' \cdot p_1, + f_3"' \cdot p_2$$
and if we assume the conditional independence of $E_1$ and $E_2$ conditioned 
on $D$, then we obtain:
$$m_{12}(\{d_1\})=e_1' \cdot f_1' \cdot p_1 +
e_1' \cdot f_3' \cdot p_1 +
e_3' \cdot f_1' \cdot p_1 $$
$$m_{12}(\{d_2\})=e_2" \cdot f_2" \cdot p_2 +
e_2" \cdot f_3" \cdot p_2 +
e_3" \cdot f_2" \cdot p_2 $$
$$m_{12}(\{d_1,d_2\})=e_3' \cdot f_3' \cdot p_1 +
e_3" \cdot f_3" \cdot p_2$$
 However, the Dempster rule of combination would result in (c - normalization 
constant):\\
$$m_1\oplus m_2(\{d_1\})=c \cdot (e_1' \cdot f_1' \cdot p_1^2 +
e_1' \cdot f_3' \cdot p_1^2         +
e_1' \cdot f_3" \cdot p_1 \cdot p_2 +
e_3' \cdot f_1' \cdot p_1^2         +
e_3" \cdot f_1' \cdot p_1 \cdot p_2)$$
$$m_1\oplus m_2(\{d_2\})=c \cdot (e_2" \cdot f_2" \cdot p_2^2 +
e_2" \cdot f_3' \cdot p_1 \cdot p_2 +
e_2" \cdot f_3" \cdot p_2^2         +
e_3' \cdot f_2" \cdot p_1 \cdot p_2 +
e_3" \cdot f_2" \cdot p_2^2        )$$
$$m_1\oplus m_2(\{d_1,d_2\})=c \cdot 
e_3' \cdot f_3' \cdot p_1^2         +
e_3" \cdot f_3" \cdot p_2^2         +
e_3' \cdot f_3" \cdot p_1 \cdot p_2 +
e_3" \cdot f_3' \cdot p_1 \cdot p_2)$$
Obviously, $Bel_{12}$ and $Bel_1\oplus Bel_2$ are not identical in general.  
We conclude that conditional independence of experts is also impossible. 
Hence no usual staatistical indeperndence assumption  is valid for the 
rough set interpretation of the DST. This fact points at where the difference 
between rough set interpretation and our interpretation lies in: in our 
 interpretation, traditional statistical independence is incorporated into 
the Dempster's scheme of combination (labelling process).

By the way, lack of correspondence between statistical independence and 
Dempster rule of combination is characteristic not only of the rough set 
interpretation, but also of most of the other ones. The 
Reader should read carefully clumsy statements of Shafer about DST 
and statistical independence in \cite{Shafer:90b}.

\subsection{General Remarks}

The Dempster-Shafer Theory exists already over two decades. Though it was 
claimed to reflect various aspects of human reasoning, it has not been widely 
used in expert systems until recently due to the high computational 
complexity. Three years ago, however, an important paper of Shenoy and Shafer 
\cite{Shenoy:90} has been published, 
along papers of other authors similar in spirit,
which meant a break-through for 
application of both bayesian and Dempster-Shafer theories in  reasoning 
systems, because it demonstrated that if joint (bayesian or DS) belief 
distribution can be decomposed in form of a belief network than it can be both 
represented in a compact manner and marginalized efficiently by local 
computations. 

This  fact  makes  them  suitable   as   alternative 
fundamentals for  representation  of  (uncertain)  knowledge in  
expert system knowledge bases \cite{Henrion:90}. 

Reasoning in bayesian belief networks has been subject of intense
research work also earlier 
 \cite{Shachter:90}, \cite{Shenoy:90}, \cite{Pearl:86},
\cite{Pearl:88}. 
There exist methods of  imposing 
various logical constraints on the probability  density  function  and  of 
calculating  marginals  not  only  of  single  variables  but  of 
complicated logical expressions over elementary statements of  the 
type X =x   (x   belonging to the domain of the variable X ) 
\cite{Pearl:88}. 
     There  exist  also  methods  determining   the 
decomposition of 
a joint probability distribution given by a sample into a 
bayesian belief network
\cite{Chow:68}, \cite{Rebane:89}, \cite{Acid:91}, \cite{Srinivas:90}.

     It  is  also  known  that  formally 
probability distributions can be treated as special cases of 
Dempster-Shafer belief distributions (with sinngleton focal points) 
\cite{Halpern:92}.
 
     However, for application of DS Belief-Functions
for representation of uncertainty in 
expert 
system knowledge bases there exist several severe  obstacles.   The 
main one  is  the  missing  frequentist  interpretation  of  the 
DS-Belief function and hence neither a comparison of the deduction 
results with experimental  data  nor  any  quantitative  nor  even 
qualitative conclusions can be drawn from results of deduction  in 
Dempster-Shafer-theory based expert systems \cite{Ma:91}.

     Numerous attempts to 
find  a frequentist  interpretation have been reported (e.g. 
\cite{Fagin:91}, \cite{Fagin:91B},
\cite{Grzymala:91},
\cite{Halpern:92}, 
\cite{Kyburg:87}, \cite{Shafer:90b},
 \cite{Skowron:93}).   But, as Smets \cite{Smets:92} states, they failed 
either trying 
 to incorporate  Dempster  rule  or  when  explaining  the  nature  of 
probability interval approximation. 
The Dempster-Shafer Theory experienced 
therefore
sharp criticism from several
authors in the past \cite{Pearl:88}, \cite{Halpern:92}.
It is suggested in those critical papers that 
the claim of DST to represent uncertainty stemming from ignorance is not 
valid.    Hence alternative rules of combination of evidence have been 
proposed. However, these rules fail to fulfill Shenoy/Shafer axioms of local 
computation \cite{Shenoy:90} and hence are not tractable in practice. These 
failures of those authors meant to us that one shall nonetheless try to find 
a meaningful frequentist interpretation of DST compatible with Dempster rule 
of combination.

     We have carefully studied several of these approaches and are 
convinced that the key for many of those  failures  (beside  those 
mentioned by Halpern in \cite{Halpern:92}) was:
(1)  treating  the  Bel-Pl 
pair as an interval approximation and
(2)  viewing combination of evidence as a process of approaching a point 
estimation. In this paper we claim 
that the most reasonable treatment of Bel's Pl's is to consider them to be 
POINT ESTIMATES of probability distribution over set-valued attributes
 (rather then Interval estimates of probability distribution over single 
valued attributes). Of course, we claim 
 also that Bel-Pl estimates by  an  interval  some  probability  density 
function but in our interpretation  that  "intrinsic"  probability 
density function is of little interest for the  user.  The combination of 
evidence represents in our interpretation manipulation of data by imposing on 
them our prejudices (rather then striving for extraction of true values).

 Under  these 
assumptions a frequentionistically meaningful interpretation to the 
Bel's  can  be  constructed,  which   remains   consistent   under 
combination  of  joint  distribution   with   "evidence",   giving 
concrete  quantitative  meaning  to  results  of   expert   system 
reasoning. Within this interpretation we were  able 
to prove the correctness of Dempster-Shafer rule. This means  that 
this frequentist interpretation  is  consistent  with  the  DS-Theory  to  the 
largest extent ever  achieved. 

\section{Conclusions}

\begin{itemize}
\item According to Smets \cite{Smets:92} there has existed no proper 
frequentist interpretation of the Dempster-Shafer theory of evidence so far.  
 \item In this paper a novel frequentist interpretation of the 
Dempster-Shafer-Theory has been found allowing for close correspondence 
between Belief and Plausibility functions and the real data.
 \item This interpretation fits completely into the framework of Bel/Pl 
definitions and into the Dempster rule of combination of independent evidence
relating for the first time in DST history this rule to plain 
statistical independence 
just overcoming difficulties of many alternative interpretations of the 
Dempster-Shafer-Theory. Hence this 
interpretation dismisses the claim of  Smets \cite{Smets:92} that such an 
interpretation cannot exist.
\item It is distinguished by the fact of postponing the moment of measuring 
 object properties behind combination of evidence leading even to dropping 
some 
costly measurements altogether.
\item The interpretation allows for subjective treatment of Bel's and Pl's as 
some approximations to unknown probability distribution of an intrinsic, but 
not accessible, attribute.
\item The introduced concept of labeled population may to some extent 
represent subjectivity in viewing probabilities.
\item This interpretation questions the common usage of the DST as a mean to 
represent and to reason with uncertainty stemming from ignorance. This view 
has been already shaken by works of Pearl \cite{Pearl:88} and Halpern and 
Fagin 
\cite{Halpern:92}. What our interpretation states clearly is that the DST 
should be viewed as a way to express unwillingness to accept objective facts 
rather than as a mean to express ignorance about them. Hence it should be 
called a theory of prejudices rather than a theory of evidence.
 \end{itemize}

Finally, I feel obliged to apologize and to say that all critical remarks 
towards 
interpretations of DST elaborated by other authors result from deviations of 
those interpretations from the formalism of the DST. I do not consider, 
 however, a deviation from DST as a crime, because modifications of DST may 
and possibly have a greater practical importance than the original theory. 
The purpose of this paper was to shed  a bit more light onto the intrinsic 
nature of  pure DST and not to call for orthodox attitudes towards DST.

\section*{Acknowledgements}

I am indebted to anonymous referees who greatly contributed to enhancement
of the quality of presentation of this paper. 

\newcommand{\ReadingsIn}{G. Shafer, J. Pearl eds: Readings in Uncertain 
Reasoning, (ISBN 1-55860-125-2, 
Morgan Kaufmann Publishers Inc., San Mateo, California, 1990)}

\end{document}